\documentclass[runningheads]{llncs}
\usepackage[dvipsnames]{xcolor}
\definecolor{Mygreen}{RGB}{89, 163, 57}
\definecolor{Myred}{RGB}{234, 51, 35}
\definecolor{Mypurple}{RGB}{155, 58, 145}
\definecolor{Myyellow}{RGB}{245, 194, 66}
\usepackage{amsmath}
\usepackage{hyperref}
\hypersetup{
    colorlinks=true,
    urlcolor=blue,
    linkcolor=black,
    citecolor=black
}
\usepackage{booktabs}
\usepackage{multirow}
\usepackage{siunitx}
\usepackage[T1]{fontenc}

\usepackage{graphicx,verbatim}
\begin{document}
\title{Fine-tuning Vision Language Models with Graph-based Knowledge for Explainable Medical Image Analysis}
\titlerunning{Tuning VLMs with graph-based knowledge}

%

\author{Chenjun Li\inst{1,4} \and
Laurin Lux\inst{2,4} \and Alexander H. Berger\inst{2,4} \and Martin J. Menten\inst{2} \and Mert R. Sabuncu\inst{1,3,4} \and Johannes C. Paetzold\inst{3,4}
 }

\authorrunning{C. Li et al.}

\institute{School of Electrical and Computer Engineering, Cornell University, Ithaca, NY 14853, USA \and
School of Computation, Information and Technology, Technical University of Munich, 80333 Munich, Germany \and
Cornell Tech, New York, NY 10044, USA \and Weill Cornell Medicine, New York, NY 10021, USA\\
\email{cl2733@cornell.edu; jpaetzold@med.cornell.edu}}

\maketitle              
\begin{abstract}

Accurate staging of Diabetic Retinopathy (DR) is essential for guiding timely interventions and preventing vision loss. However, current staging models are hardly interpretable, and most public datasets contain no clinical reasoning or interpretation beyond image-level labels. In this paper, we present a novel method that integrates graph representation learning with vision-language models (VLMs) to deliver explainable DR diagnosis. Our approach leverages optical coherence tomography angiography (OCTA) images by constructing biologically informed graphs that encode key retinal vascular features such as vessel morphology and spatial connectivity. A graph neural network (GNN) then performs DR staging while integrated gradients highlight critical nodes and edges and their individual features that drive the classification decisions. We collect this graph-based knowledge which attributes the model's prediction to physiological structures and their characteristics. We then transform this reasoning into textual descriptions for VLMs. We perform instruction-tuning with these textual descriptions and the corresponding image to train a student VLM. This final agent can classify the disease and explain its decision in a human interpretable way solely based on a single image input. Experimental evaluations on both proprietary and public datasets demonstrate that our method not only improves classification accuracy but also offers more clinically interpretable results. An expert study further demonstrates that our agent provides more accurate diagnostic explanations and enables precise localization of pathologies in OCTA images.


\end{abstract}

\section{Introduction}

Diabetic Retinopathy (DR) remains one of the primary causes of vision loss, and its early detection and staging can significantly reduce the risk of blindness \cite{lee2015epidemiology}. Early work has demonstrated the capabilities of deep learning models in the accurate prediction of DR staging on color fundus images \cite{dai2021deep,takahashi2017applying}. Optical Coherence Tomography Angiography (OCTA) is a higher resolution non-invasive imaging modality for examining retinal vasculature in fine detail. Compared to fundus images, OCTA images capture finer microvascular changes linked to DR progression. Biomarkers extracted from OCTA images, such as Blood Vessel Density (BVD) and the Foveal Avascular Zone (FAZ) area, play a critical role in evaluating the severity of DR \cite{sandhu2020automated,sun2021optical}. Methods based on these biomarkers have shown promising results: Sandhu et al. \cite{sandhu2020automated} proposed a machine learning pipeline that integrates OCT and OCTA features with clinical and demographic data to enhance the classification of non-proliferative diabetic retinopathy (NPDR). Alam et al. \cite{alam2020quantitative} developed a support vector machine classifier leveraging structural features, such as vessel tortuosity, vascular caliber, and vessel perimeter index to categorize NPDR severity levels. However, the biomarkers cannot be attributed to specific image regions, and the results are difficult to interpret or verify by clinicians.

\begin{figure}
    \centering
    \includegraphics[width=0.75\textwidth]{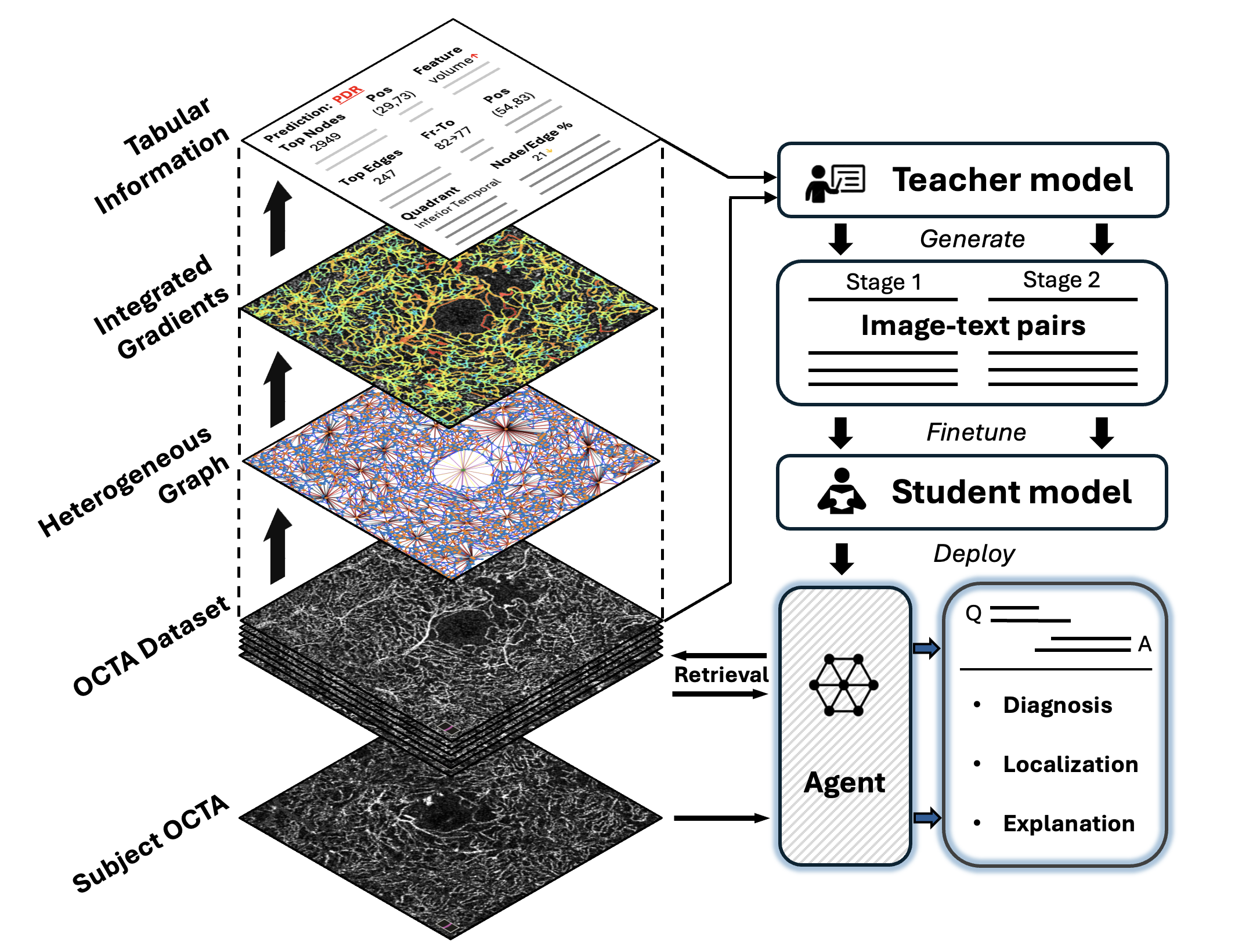}
    \caption{\textbf{Method Overview.} Biology-informed heterogeneous graphs are first constructed based on OCTA images of the DVC. Integrated gradients highlight key vascular features and regions, which are then summarized in tabular form. A teacher model uses this information to generate image-text data for fine-tuning a student model, culminating in an \textcolor{RoyalBlue}{\textbf{agent}} that provides DR diagnosis, localization of abnormalities, and explanations solely based on a given OCTA image. }
    \label{fig:overview}
\end{figure}


Developments in Vision-Language Models (VLMs) open up new possibilities for addressing the interpretability gaps in deep learning. They not only excel at generating coherent text but can also respond to specialized prompts and integrate supplementary information, whether from a retrieval-augmented pipeline \cite{xia2024rule} or through fine-tuning \cite{li2024llava,liu2023llava}. VLMs leverage massive paired datasets from the natural image domain to perform various vision-language tasks. Emerging large datasets of biomedical image-text pairs \cite{Du_RETCLIP_MICCAI2024,johnson10identified,zhang2023large} further enable the training and tuning of VLMs for domain-specific tasks. In modalities such as MRI and CT, attempts have been made to adapt VLMs for basic diagnostic reporting \cite{Sha_Fewshot_MICCAI2024,Zha_Diseaseinformed_MICCAI2024,zhang2024generalist}. Importantly, conversation offers an intuitive and natural way of exchanging knowledge \cite{fan2025ai,salminen2024using}, enabling clinicians to resolve uncertainties, seek further details, or reconcile model outputs with their own expertise. However, current methods tend to rely on generic image-text pairs that lack the granularity needed to accurately describe pathological features. Moreover, for modalities like OCTA, the scarcity of large-scale, high-quality image-text datasets further constrains the direct application of VLMs.

\subsubsection{Contribution.} In this paper, we present a novel method for training a VLM agent that enhances the interpretability of DR staging by providing a textual interface which allows clinicians to directly interact with the DR staging model. We achieve this by first integrating a GNN to capture complex spatial relationships in OCTA images of the deep vascular complex (DVC) for DR staging, and then transforming this graph-based knowledge into structured, table-formatted texts for a teacher model to generate fine-tuning data. We subsequently fine-tune VLM models with direct image inputs. The results show enhanced performance in both classification and interpretation tasks, where only OCTA images and a few lines of background knowledge are provided as prompts, marking a promising step toward models that can both classify and explain their predictions in a clinically relevant manner. In that, our method is related to current advances in reasoning language models. 

\section{Method}

\subsection{Overview}

Fig. \ref{fig:overview} outlines the major steps of the proposed method: (1) construct a heterogeneous graph from OCTA scans and train a GNN to predict DR stages; (2) employ integrated gradients to identify edges and nodes critical to the GNN’s decision; (3) consolidate these important graph elements into a structured table; (4) generate Q\&A pairs by a teacher model for vision-focused instruction tuning; and (5) during inference, the fine-tuned model receives a raw OCTA image as input and functions as an interactive diagnostic agent that combines the classification capabilities of the GNN with the rich explanatory power of VLMs.

\subsection{GNN-based Staging}
Graphs are particularly well-suited for modeling the structure of the retinal vasculature \cite{lan2024hybrid,menten2023synthetic}. Following a recent work on DR staging with GNNs \cite{Lux2025GNN}, we first construct a heterogeneous graph representation that encodes biologically relevant features of the retina. This graph consists of nodes representing vessel segments, intercapillary areas, and the FAZ, with edges capturing spatial and structural relationships. Retinal vasculature is segmented using a high-resolution method \cite{kreitner2024synthetic}, ensuring continuity for accurate graph representation. Vessel segments between bifurcation points are represented as nodes with features such as length, curvature, and radius. Intercapillary areas are detected via connected component labeling, with nodes enriched by geometric properties like area, perimeter, and eccentricity. 
 
DR staging is then treated as a graph classification task and a GNN is employed to process the constructed graph. The GNN employs multiple SAGE layers \cite{hamilton2017inductive} to perform message passing across both homogeneous and heterogeneous edges. Aggregated features are obtained via sum and max pooling, capturing both dense and sparse representations of the graph. These aggregated embeddings are then concatenated and processed through a multi-layer perceptron, which outputs predictions for DR stages as one of three discrete classes (healthy, non-proliferative DR and proliferative DR (PDR)).



\subsection{Feature and Location Attribution}
To identify the important biological features of individual nodes and edges leading to the predictions, we apply integrated gradients (IG) \cite{Lux2025GNN,sundararajan2017axiomatic}, a method that quantifies feature attributions by evaluating gradients along a path from a baseline input to the actual input. For graph-structured data, we compute IG for each node and edge to assess their contribution to the predicted outcome. For each node \(v\) and its feature \(i\), their IG is shown in equation \eqref{eqn1}.
\begin{equation}\label{eqn1}
    \mathrm{IG}_i(\mathbf{x}) = (x_{\text{input},i} - x_{\text{baseline},i}) \times \int_0^1 \frac{\partial f(\mathbf{x}_{\text{baseline}} + \alpha (\mathbf{x}_{\text{input}} - \mathbf{x}_{\text{baseline}}))}{\partial x_i} \, d\alpha,
\end{equation}

where \(f\) is the model's prediction function and \(x_i\) represents the \(i\)-th feature of node \(v\). The resulting IG scores are used to rank nodes and edges by their influence on the prediction. The advantage of these graph-based explanations is that they preserve both human interpretable features (e.g. vessel diameter) and the features' locations. 

\subsection{Data Preparation for Instruction Tuning}

\begin{figure}
    \centering
    \includegraphics[width=1.0\textwidth]{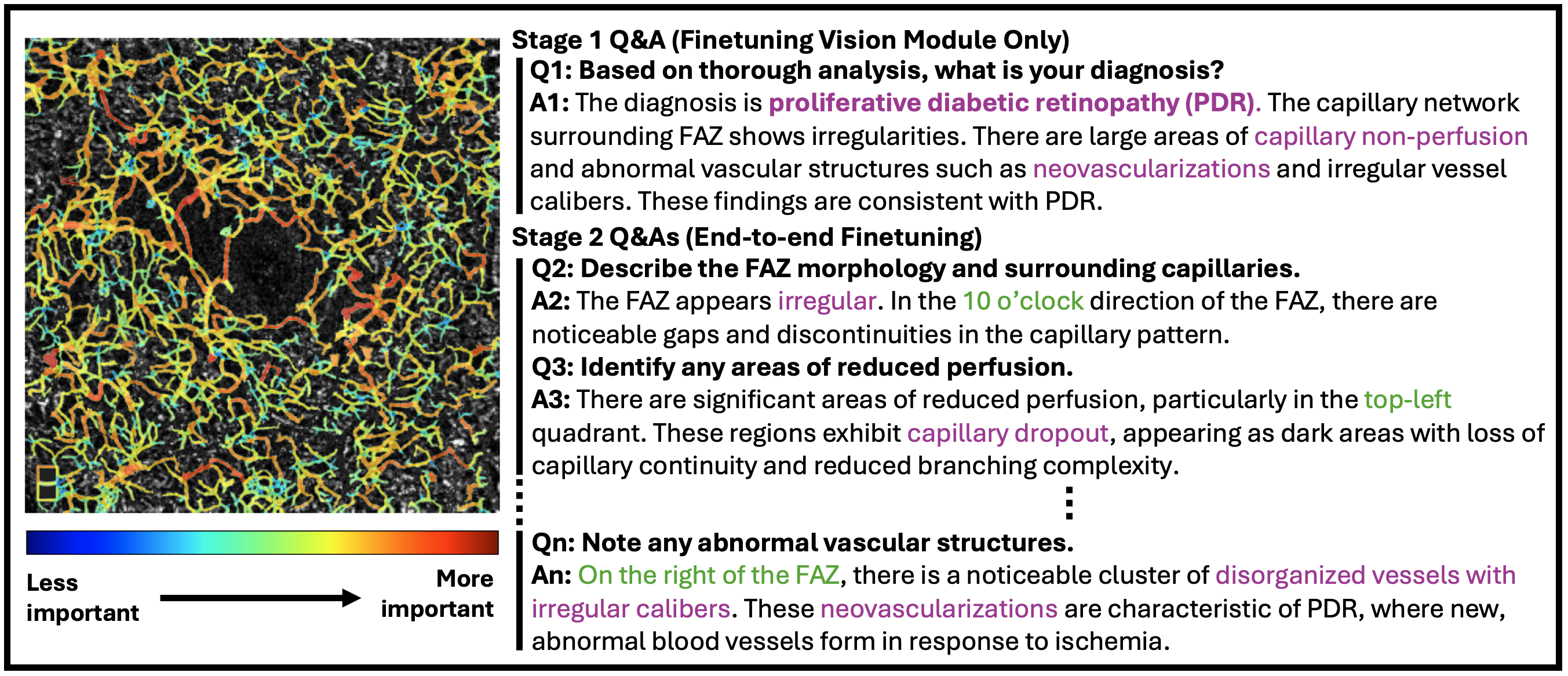}
    \caption{\textbf{Examples of Instruction Tuning Data.} Key \textcolor{Mypurple}{features} and \textcolor{Mygreen}{locations} are marked in purple and green, respectively. An overlay heatmap is provided as an example to highlight the importance of blood vessels on the OCTA image based on the IG method. During the actual training and inference, only raw OCTA images are used.}
    \label{fig:finetune_qas}
\end{figure}

The information stored on the important nodes and edges identified by IG is compiled into a structured table. Specifically, for each graph, the table includes: (1) graph-level information: densities of nodes and edges in each of the four quadrants. (2) Node-level information: Node ID, importance score, spatial location, and the most important features such as vessel diameter and roundness. (3) Edge-level information: Edge ID, importance score, connected node pairs, spatial location, and most important features. The ground truth label of the image and the classification probabilities generated by the GNN are also provided to guide the teacher model. 

The structured tables and OCTA images are then used to generate question-answer pairs for two-stage instruction tuning. OpenAI o1 \cite{openai_o1} is used as a teacher to create conversational datasets that simulate clinical interactions. Example Q\&A pairs generated by the teacher model are shown in Fig. \ref{fig:finetune_qas}, the questions in stage 1 focus on overall diagnosis, while stage 2 incorporates more detailed, location-specific questions that probe deeper into retinal morphology, including specific quadrant abnormalities and distinctive vascular patterns.

\subsection{VLM Prompting and Tuning}
We adapt our method to lightweight (open source) models and state-of-the-art (closed) commercial models. For lightweight models such as Llama-3.2-11B-Vision \cite{touvron2023llama} and Qwen 2.5-VL 7b \cite{yang2024qwen2}, we follow similar steps as described in \cite{li2024llava,liu2023llava} to fine-tune the VLMs. Specifically, in the first stage, we freeze the language model and align the model based on image-classification pairs. In the second stage, we fine-tune the model end-to-end with stage-2 Q\&A pairs.

For larger commercial models, we follow the standard procedures in their APIs to perform supervised fine-tuning. Since these models allow extended context windows, we employ a retrieval process for each set of Q\&A data for an image. This process adds cases with similar graph features along with their labels and images as context, helping the models to better understand the graph-based knowledge. Specifically, we compute an eight-dimensional distribution vector that quantifies the number of nodes and edges located in each of the four quadrants of the image. During inference, we calculate the Euclidean distance between the distribution vector of the test sample and those of the training samples, and retrieve the top three most similar cases. These similar cases are also incorporated into the prompt for the VLMs. 

\section{Experiments}

We conduct comprehensive experiments with 4 state-of-the-art VLMs to evaluate our method's quantitative and qualitative performance in DR staging and explanation. Code is available at \href{https://github.com/chenjun-li/GFT}{https://github.com/chenjun-li/GFT}.

\noindent \textbf{Dataset.} \space We test the models on both a proprietary dataset and a public dataset. The proprietary dataset consists of 1264 high-quality OCTA images of the DVC and is used to train and validate the GNN. Each of these images has a resolution of $304\times304$ pixels and is assigned one of the three DR staging labels (Healthy, PDR, and NPDR). We divide the proprietary dataset into six splits, separate one as a fixed, never-seen test set, and perform five-fold cross-validation training on the other five splits. We use the public OCTA \cite{li2024octa} dataset as an additional test set. We select 189 images that are either healthy or DR (160 Healthy, 29 DR) and disregard images with other diseases. During inference, we pool PDR and NPDR to a single label.

\noindent \textbf{Fine-tuning.} \space  In the first stage of fine-tuning lightweight models, we use $844$ pairs of Q\&As with only questions asking about the staging classification, and answers explaining the key morphological features that lead to the prediction. In the second stage, $844\times30$ pairs of Q\&As with questions asking about DR diagnosis and abnormalities in specific regions are used. We fine-tune both the vision module and the language module using LoRA \cite{hu2021lora} on three NVIDIA RTX A6000 GPUs. For larger commercial models, we perform fine-tuning using their APIs and default settings.

\noindent \textbf{Evaluation Metrics.} \space For classification performance, we evaluate all models using balanced accuracy, precision, and recall rate. For explanation performance, we first follow previous works \cite{li2024llava,liu2023llava} and use the ground-truth diagnosis and graph-based knowledge to prompt the teacher model to generate a set of responses as standards, and then ask it to compare with the candidate models' responses and give scores (0-100) based on the quality of the explanations. Furthermore, we present 48 responses of each model to two ophthalmology experts. These 48 responses are generated by randomly shuffling the outputs from 6 models across 8 samples. The experts then rank and assign scores to the responses based on three criteria: overall accuracy, correct localizations, and helpfulness. A quadrant-based system is used to verify localizations correspondence. A region is marked as correct only when both experts agree. The ratings for each model are then averaged to obtain the final scores.

\begin{table}
\centering
\footnotesize
\caption{\textbf{DR Staging Classification Performance.} The proposed GFT models consistently outperform FT and BS models.}
\label{tab:dr_performance}
\begin{tabular}{lccccccc}
\toprule
\textbf{Model} & \shortstack{Bal.\\Acc.} & \shortstack{Prec.\\Heal.} & \shortstack{Rec.\\Heal.} & \shortstack{Prec.\\PDR} & \shortstack{Rec.\\PDR} & \shortstack{Prec.\\NPDR} & \shortstack{Rec.\\NPDR} \\
\midrule
BS-GPT-4o         & 0.360 & 0.683 & 0.502 & 0.067 & 0.125 & 0.185 & 0.453 \\
BS-Qwen-VL-max    & 0.303 & 0.792 & 0.381 & 0.071 & 0.188 & 0.191 & 0.341 \\
BS-Llama 3.2 11b  & 0.237 & 0.742 & 0.331 & 0.071 & 0.188 & 0.111 & 0.192 \\
BS-Qwen-2.5VL 7b  & 0.285 & 0.731 & 0.412 & 0.073 & 0.251 & 0.121 & 0.193 \\
\midrule
FT-GPT-4o              & 0.450 & 0.819 & 0.443 & 0.118 & 0.501 & 0.233 & 0.407 \\
FT-Qwen-VL-max         & 0.474 & 0.917 & 0.547 & 0.334 & 0.063 & 0.235 & \textbf{0.813} \\
FT-Llama 3.2 11b       & 0.652 & 0.930 & 0.919 & 0.702 & 0.388 & 0.517 & 0.650 \\
FT-Qwen-2.5VL 7b       & 0.569 & 0.897 & \textbf{0.957} & 0.462 & 0.375 & 0.500 & 0.375 \\
    \midrule
GFT-GPT-4o         & 0.569 & 0.927 & 0.858 & 0.254 & 0.438 & 0.294 & 0.594 \\
GFT-Qwen-VL-max    & 0.574 & 0.883 & 0.913 & 0.382 & 0.375 & 0.352 & 0.656 \\
GFT-Llama 3.2 11b     & \textbf{0.678} & \textbf{0.935} & 0.921 & \textbf{0.712} & \textbf{0.548} & \textbf{0.523} & 0.568 \\
GFT-Qwen-2.5VL 7b     & 0.613 & 0.925 & 0.902 & 0.556 & 0.313 & 0.465 & 0.625 \\
\bottomrule
\end{tabular}
\end{table}

\section{Results and Discussion}

\subsection{Diabetic Retinopathy Staging Results}

Table~\ref{tab:dr_performance} presents the DR staging performance on our proprietary dataset. Our graph-knowledge-fine-tuned (GFT) models consistently outperform the baseline vision-language models (BS) and standard fine-tuning (FT) approaches. GFT-Llama 3.2 11b achieves a balanced accuracy of 0.678, only slightly lower than the specialized GNN's performance, reported as 0.689 on the identical test set in \cite{Lux2025GNN}. Across architectures, GFT brings an average improvement of 14.8\% in balanced accuracy to FT, demonstrating the effectiveness of integrating graph. The confidence intervals from the 5-fold cross-validation are: balanced accuracy $\pm 10.13\%$, precision $\pm 5.67\%$, and recall $\pm 4.47\%$, respectively.

Table~\ref{tab:octa500_results} demonstrates the cross-dataset generalization on OCTA-500. Here, the GFT-Llama 3.2 11b model achieves a balanced accuracy of 0.842 for binary DR detection, which is comparable to the performance of a specialized GNN (0.893) and significantly better than ResNet (0.586). Compared to commercial models, the open-source models offer superior classification performance in our experiments, possibly because our end-to-end training allows for more flexible and fine-grained parameter adjustments. These results show that the graph-knowledge integration strategy not only improves performance on proprietary data but also facilitates effective knowledge transfer to public datasets.

\subsection{Quality of Explanation}

\begin{figure}
    \centering
    \includegraphics[width=1.0\textwidth]{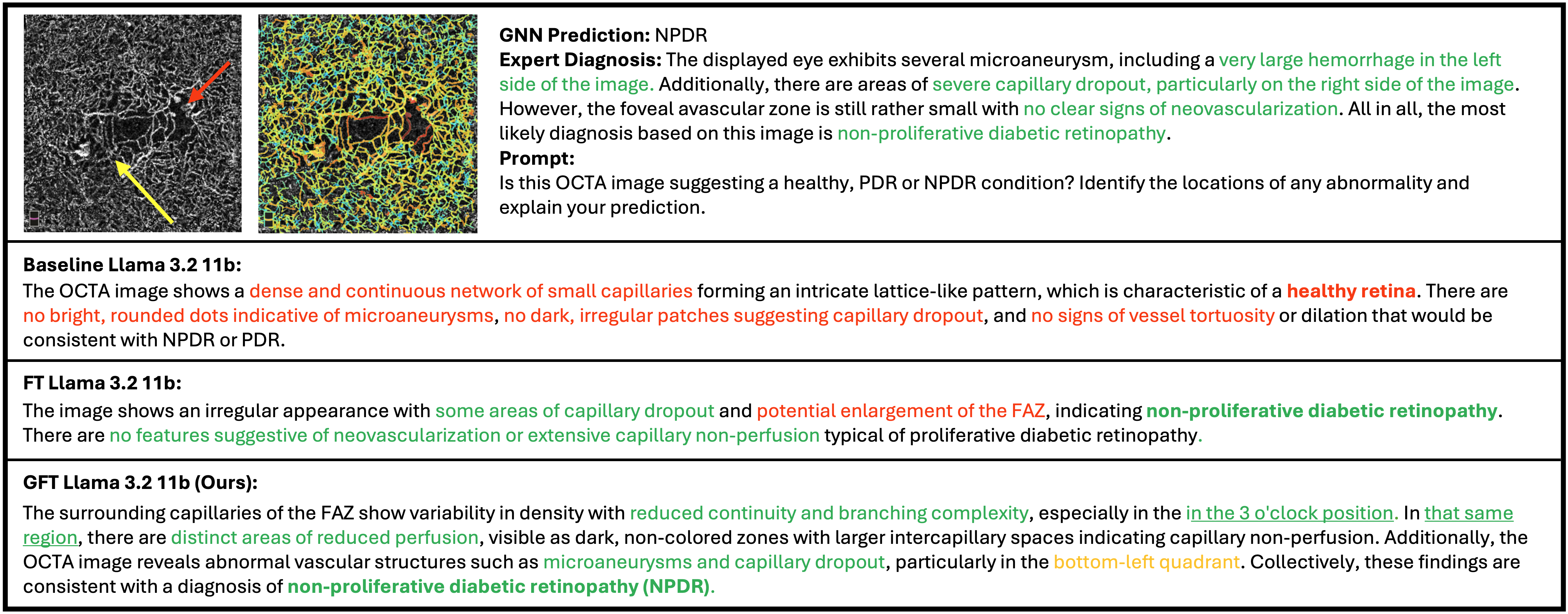}
    \caption{\textbf{Comparison of Interpretability.} \textcolor{Mygreen}{True}, \textcolor{Myyellow}{partially true} and \textcolor{Myred}{false} information is marked in green, yellow and red, respectively. An expert diagnosis is also provided. BS model makes incorrect predictions with inaccurate explanations, FT model can make correct predictions but their explanations are insufficient. GFT model can precisely identify and locate specific retinal abnormalities, closely matching expert assessment.}
    \label{fig:qtest}
\end{figure}

\begin{table}
\centering
\begin{minipage}[t]{0.45\textwidth}
\centering
\caption{\textbf{Classification Performance on OCTA-500.} We present results for BS and GFT, and for traditional image, biomarker and GNN baselines. Among VLMs, our GFTs generalize better to the unseen dataset.}
\label{tab:octa500_results}
\begin{tabular}{lccc}
\toprule
\textbf{Model} & \shortstack{\textbf{Bal.}\\\textbf{Acc.}} & \shortstack{\textbf{F1}\\\textbf{Heal.}} & \shortstack{\textbf{F1}\\\textbf{DR}} \\
\midrule
BS-GPT-4o             & 0.681   & 0.435   & \textbf{0.874}   \\
BS-Llama 3.2      & 0.551   & 0.807   & 0.256   \\
GFT-GPT-4o            & 0.759   & 0.958   & 0.682   \\
GFT-Llama 3.2     & \textbf{0.842}   & \textbf{0.969}   & 0.800 \\
\midrule
ResNet 18             & 0.586   & 0.930   & 0.294   \\
Biomarkers            & 0.821   & 0.963   & 0.760   \\
GNN \cite{Lux2025GNN} & \textbf{0.893}   & \textbf{0.978}   & \textbf{0.868}   \\
\bottomrule
\end{tabular}
\end{minipage}%
\hfill
\begin{minipage}[t]{0.53\textwidth}
\centering
\caption{\textbf{Explanation Quality.} \textit{Tch.} denotes teacher model scores, \textit{Exp.} represents expert ratings, \textit{Loc.} indicates correct region localizations across all responses, and \textit{Avg.} is the mean of teacher and expert scores. Only the proposed GFT models can provide explanations with correct localizations.}
\label{tab:qual}
\begin{tabular}{lc|cc|c}
\toprule
\textbf{Model}  & \shortstack{\textbf{Tch.}} & \shortstack{\textbf{Exp.}} & \shortstack{\textbf{Loc.}} & \textbf{Avg.} \\
\midrule
BS-GPT-4o            & 37.18  & 63.91  & 0  & 51.25 \\
BS-Llama 3.2    & 20.65   &  46.41   & 0  & 34.08\\
\midrule
FT-GPT-4o            & 48.31   & 76.25   &  0  & 62.60 \\
FT-Llama 3.2     & 52.83   & 59.31   & 0  & 58.61\\
\midrule
GFT-GPT-4o        & 62.40   &  \textbf{94.69} & \textbf{13}   &  \textbf{78.55}\\
GFT-Llama 3.2   & \textbf{68.12}  & 81.87  & 7  & 74.99 \\
\bottomrule
\end{tabular}
\end{minipage}
\end{table}

Table~\ref{tab:qual} presents a quantitative evaluation of explanation quality across different model configurations. We assess explanation quality through both automated metrics (teacher model evaluation) and human expert review. The teacher model scores reflect alignment with ground truth explanations generated from the graph-based knowledge, while expert scores evaluate clinical relevance and accuracy. The \textit{Loc.} metric quantifies a model's ability to correctly identify specific retinal regions containing pathological features across all test cases. The inter-rater weighted agreement is $\kappa = 0.83$, indicating almost perfect agreement.

The results demonstrate improvements in explanation quality through graph-based knowledge integration. GFT models consistently outperform BS and FT models across all metrics. GFT is the only model that can \textit{locate pathological changes} in the images, while all others fail to provide any region-specific explanations. Fig. \ref{fig:qtest} provides a qualitative comparison demonstrating how graph-based knowledge integration enables models to provide more clinically relevant explanations that focus on specific vascular abnormalities. Two extra examples, including exemplary interaction with the model are provided in the supplement as a video.

\section{Conclusion}
In this paper, we introduce a novel method that integrates graph-based knowledge with VLMs to facilitate end-to-end explainable diabetic retinopathy diagnosis. By constructing biologically informed heterogeneous graphs from OCTA images and applying integrated gradients for feature attribution, our approach translates complex vascular patterns into structured textual descriptions for effective instruction tuning. Experimental results on both proprietary and public data demonstrate that our method not only improves DR staging accuracy, but also generates more clinically interpretable explanations. Future work could explore the potential of the interaction in a clinical setting, and the use of synthesized images to augment the training data to further enhance model robustness.

\begin{credits}

\subsubsection{\discintname} The authors have no competing interests to declare that are relevant to the content of this article.

\end{credits}

%
%
%
\bibliographystyle{splncs04}
\bibliography{myrefs}

%




\end{document}